\def\year{2018}\relax
\documentclass[letterpaper]{article}
\usepackage{aaai18}
\usepackage{times}
\usepackage{helvet}
\usepackage{courier}

\usepackage{array}

\usepackage[utf8]{inputenc} 
\usepackage[T1]{fontenc}    
\usepackage{hyperref,color}       
\usepackage{url}            
\usepackage{booktabs}       
\usepackage{amsfonts}       
\usepackage{nicefrac}       
\usepackage{microtype}      

\usepackage{array}
\usepackage{amssymb,amsmath}
\usepackage{graphicx} 
\usepackage{subfigure} 
\usepackage{multirow}

\usepackage{algorithm}
\usepackage{algpseudocode}
\usepackage{url}

\usepackage{kantlipsum}

\usepackage{amsmath,color}
\usepackage{multirow}
\usepackage{wrapfig}
\usepackage{tabularx}
\usepackage{mydefs}
\usepackage[misc]{ifsym}

\newcommand{\ie}{\textit{i.e.}}
\newcommand{\RN}[1]{%
	\textup{\lowercase\expandafter{\it \romannumeral#1}}%
}

\begin{document}
	\title{Zero-Shot Learning via Class-Conditioned Deep Generative Models}
	\author{Wenlin Wang$^{1}${\thanks{Corresponding authors}}, Yunchen Pu$^{1}$, Vinay Kumar Verma$^{3}$, Kai Fan$^{2}$,  Yizhe Zhang$^{2}$ \\ {\Large\bf{Changyou Chen$^{4}$, Piyush Rai$^{3 *}$, Lawrence Carin$^{1}$}} \\
		$^{1}$Department of Electrical and Computer Engineering, Duke University\\
		$^{2}$Compuational Biology and Bioinformatics, Duke University\\
		$^{3}$Department of Computer Science and Engineering, IIT Kanpur, India \\
		$^{4}$Department of Computer Science and Engineering, SUNY at Buffalo\\
			{\footnotesize
				{{\{ww107, yp42, kf96, yz196, lcarin\}@duke.edu}},
				{{\{vkverma, piyush\}@cse.iitk.ac.in}},
				{{cchangyou@gmail.com}}
			}
	}

\maketitle
\begin{abstract}
	We present a deep generative model for Zero-Shot Learning (ZSL). Unlike most existing methods for this problem, that represent each class as a \emph{point} (via a semantic embedding), we represent each seen/unseen class using a class-specific \emph{latent-space distribution}, conditioned on class attributes. We use these latent-space distributions as a prior for a \emph{supervised} variational autoencoder (VAE), which also facilitates learning highly discriminative feature representations for the inputs. The entire framework is learned end-to-end using only the seen-class training data. At test time, the label for an unseen-class test input is the class that maximizes the VAE lower bound. 
	We further extend the model to a (\RN{1}) semi-supervised/transductive setting by leveraging unlabeled unseen-class data via an \emph{unsupervised} learning module, and (\RN{2}) few-shot learning where we also have a small number of labeled inputs from the unseen classes. We compare our model with several state-of-the-art methods through a comprehensive set of experiments on a variety of benchmark data sets.
\end{abstract}

\section{Introduction}
A goal of autonomous learning systems is the ability to learn new concepts even when the amount of supervision for such concepts is scarce or non-existent. This is a task that humans are able to perform effortlessly. Endowing machines with similar capability, however, has been challenging. Although machine learning and deep learning algorithms can learn reliable classification rules when supplied with abundant labeled training examples per class, their generalization ability remains poor for classes that are not well-represented (or not present) in the training data. This limitation has led to significant recent interest in zero-shot learning (ZSL) and one-shot/few-shot learning~\cite{socher2013zero,lampert2014attribute,fei2006one,lake2015human,vinyals2016matching,ravi2017optimization}. We provide a more detailed overview of existing work on these methods in the Related Work section.

In order to generalize to previously unseen classes with no labeled training data, a common assumption is the availability of side information about the classes. The side information is usually provided in the form of class attributes (human-provided or learned from external sources such as Wikipedia) representing semantic information about the classes, or in the form of the similarities of the unseen classes with each of the seen classes. The side information can then be leveraged to design learning algorithms~\cite{socher2013zero} that try to transfer knowledge from the seen classes to unseen classes (by linking corresponding attributes). 

Although this approach has shown promise, it has several limitations. For example, most of the existing ZSL methods assume that each class is represented as a fixed point (e.g., an embedding) in some semantic space, which does not adequately account for intra-class variability~\cite{akata2015evaluation,mukherjee2016gaussian}. Another limitation of most existing methods is that they usually lack a proper generative model~\cite{kingma2014auto,rezende2014stochastic,kingma2014semi} of the data. Having a generative model has several advantages~\cite{kingma2014auto,rezende2014stochastic,kingma2014semi}, such as unraveling the complex structure in the data by learning expressive feature representations and the ability to seamlessly integrate unlabeled data, leading to a transductive/semi-supervised estimation procedure. This, in the context of ZSL, may be especially useful when the amount of labeled data for the seen classes is small, but otherwise there may be plenty of unlabeled data from the seen/unseen classes.

\begin{figure*}
	\begin{center}
		\includegraphics[width=\textwidth]{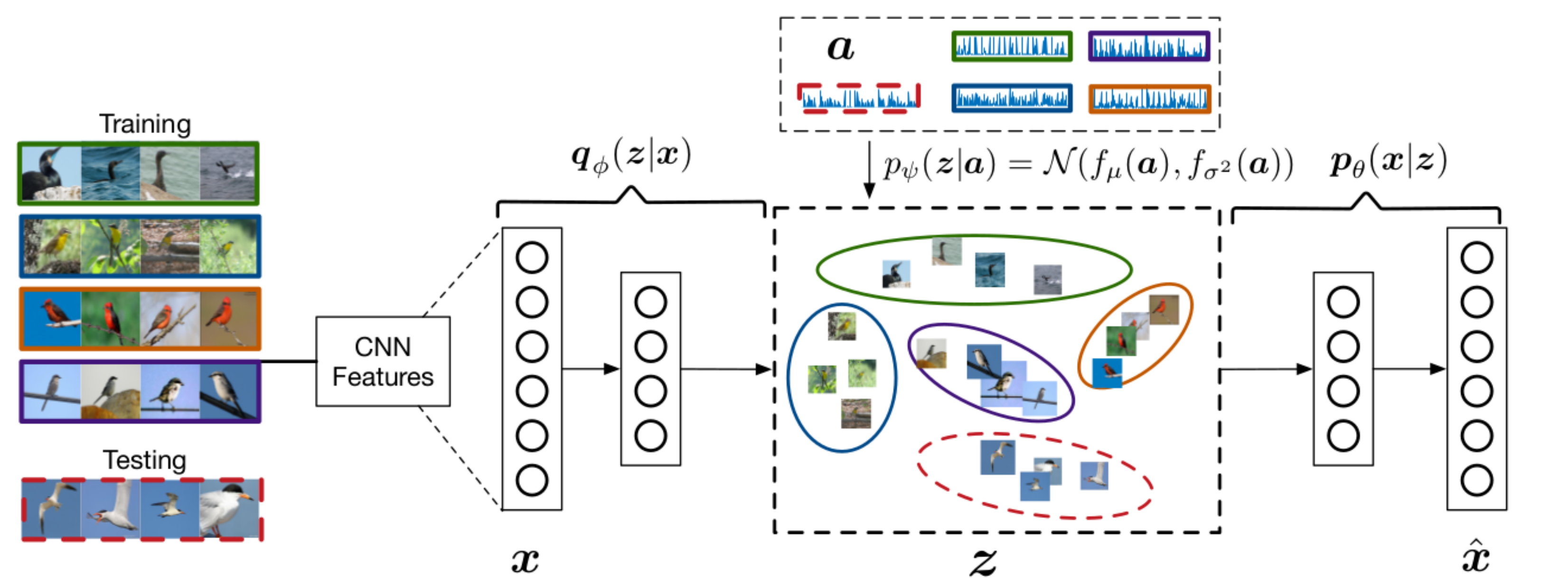}
	\end{center}
	\vspace{-1.5em}
	\caption{\small{A diagram of our basic model; only the training stage is shown here. In the above figure, $\av \in \mathbb{R}^M$ denotes the class attribute vector (given for training data, inferred for test data). Red-dotted rectangle/ellipse correspond to the unseen classes. Note: The CNN module is not part of our framework and is only used as an initial feature extractor, on top of which the rest of our model is built. The CNN can be replaced by any feature extractor depending on the data type}}
	\label{fig:idea}
	\vspace{-1em}
\end{figure*}

Motivated by these desiderata, we design a deep generative model for the ZSL problem. Our model (summarized in Figure~\ref{fig:idea}) learns a set of \emph{attribute-specific} latent space distributions (modeled by Gaussians), whose parameters are outputs of a trainable deep neural network (defined by $p_\psi$ in Figure~\ref{fig:idea}). The attribute vector is denoted as $\av$, and is assumed given for each training image, and it is inferred for test images. The class label is linked to the attributes, and therefore by inferring attributes of a test image, there is an opportunity to recognize classes at test time that were not seen when training. These latent-space distributions serve as a prior for a variational autoencoder (VAE)~\cite{kingma2014auto} model (defined by a decoder $p_\theta$ and an encoder $q_\phi$ in Figure~\ref{fig:idea}). This combination further helps the VAE to learn discriminative feature representations for the inputs. Moreover, the generative aspect also facilitates extending our model to semi-supervised/transductive settings (omitted in Figure~\ref{fig:idea} for brevity, but discussed in detail in the Transductive ZSL section) using a deep \emph{unsupervised} learning module. All the parameters defining the model, including the deep neural-network parameters $\psi$ and the VAE decoder and encoder parameters $\theta, \phi$, are learned end-to-end, using only the seen-class labeled data (and, optionally, the available unlabeled data when using the semi-supervised/transductive setting). 

Once the model has been trained, it can be used in the ZSL setting as follows. Assume that there are classes we wish to identify at test time that have not been seen when training. While we have not seen images before from such classes, it is assumed that we know the attributes of these previously unseen classes. The latent space distributions $p_\psi(\zv|\av)$ for all the unseen classes (Figure~\ref{fig:idea}, best seen in color, shows this distribution for one such unseen class using a red-dotted ellipse) are inferred by conditioning on the respective class attribute vectors $\av$ (including attribute vectors for classes not seen when training). Given a test input $\xv_*$ from some unseen class, the associated class attributes $\av_*$ are predicted by first mapping $\xv_*$ to the latent space via the VAE recognition model $q_\phi(\zv_*|\xv_*)$, and then finding $\av_*$ that maximizes the VAE lower bound. The test image is assigned a class label $y_*$ linked with $\av_*$.
This is equivalent to finding the class latent distribution $p_\psi$ that has the smallest KL divergence w.r.t. the variational distribution $q_\phi(\zv_*|\xv_*)$.

\section{Variational Autoencoder}

The variational autoencoder (VAE) is a deep generative model~\cite{kingma2014auto,rezende2014stochastic}, capable of learning complex density models for data via latent variables. Given a nonlinear generative model $p_{\theta}(\xv|\zv)$ with input $\xv \in \mathbb{R}^D$ and associated latent variable $\zv \in \mathbb{R}^L$ drawn from a prior distribution $p_0(\zv)$, the goal of the VAE is to use a recognition model $q_{\phi}(\zv|\xv)$ (also called an inference network) to approximate the posterior distribution of the latent variables, i.e., $p_{\theta}(\zv|\xv)$, by maximizing the following variational lower bound
\begin{align*}
	\Lcal_{\theta,\phi}^{\mbox{v}}(\xv) = \E_{q_\phi(\zv|\xv)}[\log p_\theta(\xv|\zv)] - \mbox{KL}(q_\phi(\zv|\xv) || p_0(\zv))~.
\end{align*}
Typically, $q_{\phi}(\zv|\xv)$ is defined as an isotropic normal distribution with its mean and standard deviation the output of a deep neural network, which takes $\xv$ as input. After learning the VAE, a probabilistic ``encoding'' $\zv$ for the input $\xv$ can be generated efficiently from the recognition model $q_{\phi}(\zv|\xv)$.

We leverage the flexibility of the VAE to design a \emph{structured}, supervised VAE that allows us to incorporate class-specific information (given in the form of class attribute vectors $\av$). This enables one to learn a deep generative model that can be used to predict the labels for examples from classes that were not seen at training time (by linking inferred attributes to associated labels, even labels not seen when training).


\section{Deep Generative Model for ZSL}
We consider two settings for ZSL learning: inductive and transductive. In the standard inductive setting, during training, we only assume access to labeled data from the seen classes. In the transductive setting~\cite{kodirov2015unsupervised}, we also assume access to the \emph{unlabeled} test inputs from the unseen classes. In what follows, under the \emph{Inductive ZSL} section, we first describe our deep generative model for the inductive setting. Then, in the \emph{Transductive ZSL} section, we extend this model for the transductive setting, in which we incorporate an \emph{unsupervised} deep embedding module to help leverage the \emph{unlabeled} inputs from the unseen classes. Both of our models are built on top of a variational autoencoder~\cite{kingma2014auto,rezende2014stochastic}. However, unlike the standard VAE~\cite{kingma2014auto,rezende2014stochastic}, our framework leverages attribute-specific latent space distributions which act as the prior (Figure~\ref{fig:idea}) on the latent codes of the inputs. This enables us to adapt the VAE framework for the problem of ZSL.

\vspace{-1em}
\paragraph{Notation} In the ZSL setting, we assume there are $S$ seen classes and $U$ unseen classes. For each seen/unseen class, we are given side information, in the form of $M$-dimensional class-attribute vectors~\cite{socher2013zero}. The side information is leveraged for ZSL. We collectively denote the attribute vectors of all the classes using a matrix $\Amat \in \R^{M\times (S+U)}$. During training, images are available only for the seen classes, and the labeled data are denoted $\Dcal_s = \{(\xv_n, \av_n)\}_{n=1}^{N}$, where $\xv_n \in \mathbb{R}^D$ and $\av_n = \Amat_{y_n}$, $\Amat_{y_n}\in \mathbb{R}^M$ denotes the $y_n^{th}$ column of $\Amat$ and $y_n \in \{1,\dots, S\}$ is the corresponding label for $\xv_n$. The remaining classes, indexed as $\{S+1, \dots,S+U\}$, represent the unseen classes (while we know the $U$ associated attribute vectors, at training we have no corresponding images available). Note that each class has a unique associated attribute vector, and we infer unseen classes/labels by inferring the attributes at test, and linking them to a label.

\subsection{Inductive ZSL}
\label{sec:ind}

We model the data $\{\xv_n\}_{n=1}^N$ using a VAE-based deep generative model, defined by a decoder $p_\theta(\xv_n|\zv_n)$ and an encoder $q_\phi(\zv_n|\xv_n)$. As in the standard VAE, the decoder $p_{\theta}(\xv_n|\zv_n)$ represents the generative model for the inputs $\xv_n$, and $\theta$ represents the parameters of the deep neural network that define the decoder. Likewise, the encoder $q_\phi(\zv_n|\xv_n)$ is the VAE \emph{recognition model}, and $\phi$ represents the parameters of the deep neural network that define the encoder. 

However, in contrast to the standard VAE prior that assumes each latent embedding $\zv_n$ to be drawn from the same latent Gaussian (e.g., $p_{\psi}(\zv_n) = \Ncal(0,\Imat)$), we assume each $\zv_n$ to be drawn from a \emph{attribute-specific} latent Gaussian, $p_{\psi}(\zv_n|\av_n) = \Ncal(\muv(\av_n), \Sigmamat(\av_n))$, where 
\beq 
\muv(\av_n) = f_\mu(\av_n),~~ \Sigmamat(\av_n) = \text{diag}(\exp{(f_\sigma(\av_n))})
\label{eq:meancov}
\eeq
where we assume $f_\mu(\cdot)$ and $f_\sigma(\cdot)$ to be linear functions, \ie, $f_\mu(\av_n) = \Wmat_\mu \av_n$ and $f_\sigma(\av_n) = \Wmat_\sigma \av_n$; $\Wmat_\mu$ and $\Wmat_\sigma$ are learned parameters. 
One may also consider  $f_\mu(\cdot)$ and $f_\sigma(\cdot)$ to be a deep neural network; this added complexity was not found necessary for the experiments considered.
Note that once $\Wmat_\mu$ and $\Wmat_\sigma$ are learned, the parameters $\{\muv(\av), \Sigmamat(\av)\}$ of the latent Gaussians of unseen classes $c=S+1,\ldots,S+U$ can be obtained by plugging in their associated class attribute vectors $\{\Amat_c\}_{c=S+1}^{S+U}$, and inferring which provides a better fit to the data. 

Given the class-specific priors $p_{\psi}(\zv_n|\av_n)$ on the latent code $\zv_n$ of each input, we can define the following variational lower bound for our VAE based model (we omit the subscript $n$ for simplicity)
\begin{small}
	\begin{align}
		\vspace{-1em}
		\Lcal_{\theta,\phi,\psi}(\xv, \av) = \E_{q_\phi(\zv|\xv)}[\log p_\theta(\xv|\zv)] - \mbox{KL}(q_\phi(\zv|\xv) || p_\psi(\zv|\av))\label{eq:elbo}
		\vspace{-0.5em}
	\end{align}
\end{small}

\vspace{-1em}
\noindent\textbf{Margin Regularizer} The objective in (\ref{eq:elbo}) naturally encourages the inferred variational distribution $q_\phi(\zv|\xv)$ to be close to the class-specific latent space distribution $p_\psi(\zv|\av)$. However, since our goal is classification, we augment this objective with a \emph{maximum-margin} criterion that promotes $q_\phi(\zv|\xv)$ to be as far away as possible from all other class-specific latent space distributions $p_{\psi}(\zv|\Amat_c)$, $\Amat_c \neq \av$. To this end, we replace the $- \mbox{KL}(q_\phi(\zv|\xv) || p_\psi(\zv|\av))$ term in our original VAE objective (\ref{eq:elbo}) by $-[\mbox{KL}(q_\phi(\zv|\xv) || p_\psi(\zv|\av)) - R^*]$ where  ``margin regularizer'' term $R^*$ is defined as the minimum of the KL divergence between $q_\phi(\zv|\xv)$ and all other class-specific latent space distributions:
{\small
\begin{align}
	R^* &= \min_{c: c\in \{1.., y-1, y+1,..,S\}} \{\mbox{KL}(q_\phi(\zv|\xv)||p_\psi(\zv|\Amat_c))\} ~\nonumber\\
	&=  -\max_{c: c\in \{1.., y-1, y+1,..,S\}} \{-\mbox{KL}(q_\phi(\zv|\xv)||p_\psi(\zv|\Amat_c))\}
\end{align}}

Intuitively, the regularizer $-[\mbox{KL}(q_\phi(\zv|\xv) || p_\psi(\zv|\av)) - R^*]$ encourages the true class and the \emph{next best} class to be separated maximally. However, since $R^*$ is non-differentiable, making the objective difficult to optimize in practice, we approximate $R^*$ by the following surrogate:
\vspace{-0.75em}
\begin{align}
	R = -\log\sum_{c=1}^S \exp(-\mbox{KL}(q_\phi(\zv|\xv)||p_\psi(\zv|\Amat_c)))
\end{align}

\noindent It can be easily shown that
\begin{align}
	R^*  \leq R \leq R^* + \log S
\end{align}

Therefore when we maximize $R$, it is equivalent to maximizing a lower bound on $R^*$. Finally, we optimize the variational lower bound together with the margin regularizer as 
{\small
\begin{align}
	\hat{\Lcal}_{\theta,\phi,\psi}(\xv, \av) &= \E_{q_\phi(\zv|\xv)}[\log p_\theta(\xv|\zv)] - \mbox{KL}(q_\phi(\zv|\xv) || p_\psi(\zv|\av)) \nonumber \\
	&\underbrace{-\lambda\log\sum_{c=1}^S \exp(-\mbox{KL}(q_\phi(\zv|\xv)||p_\psi(\zv|\Amat_c)))}_{R}\label{eq:loss}
\end{align}}

\vspace{-1em}
\noindent where $\lambda$ is a hyper-parameter controlling the extent of regularization. We train the model using the \textit{seen-class} labeled examples $\Dcal_s = \{(\xv_n, \av_n)\}_{n=1}^{N}$ and learn the parameters $(\theta, \phi, \psi)$ by maximizing the objective in (\ref{eq:loss}). Once the model parameters have been learned, the label for a new input $\hat{\xv}$ from an \emph{unseen} class can be predicted by first predicting its latent embedding $\hat{\zv}$ using the VAE recognition model, and then finding the ``best'' label by solving
\begin{align}
	\hat{y} &= \arg \max_{y \in \Ycal_u} \Lcal_{\theta,\phi,\psi}(\hat{\xv}, \Amat_y) \nonumber \\
	&= \arg \min_{y \in \Ycal_u} \mbox{KL}(q_\phi(\hat{\zv}|\hat{\xv}) || p_\psi(\hat{\zv}|\Amat_y)) 
\end{align}
where $\Ycal_u = \{S+1,\ldots,S+U\}$ denotes the set of unseen classes. Intuitively, the prediction rule assigns $\hat{\xv}$ to that unseen class whose class-specific latent space distribution $p_\psi(\hat{\zv}|\av)$ is most similar to the VAE posterior distribution $q_\phi(\hat{\zv}|\hat{\xv})$ of the latent embeddings. Unlike the prediction rule of most ZSL algorithms that are based on simple Euclidean distance calculations of a point embedding to a set of ``class prototypes''~\cite{socher2013zero}, our prediction rule naturally takes into account the possible \emph{multi-modal} nature of the class distributions and therefore is expected to result in better prediction, especially when there is a considerable amount of intra-class variability in the data.

\subsection{Transductive ZSL}\label{sec:trans}

We now present an extension of the model for the \emph{transductive} ZSL setting~\cite{kodirov2015unsupervised}, which assumes that the test inputs $\{\hat{\xv}_i\}_{i=1}^{N^\prime}$  from the unseen classes are also available while training the model. Note that, for the inductive ZSL setting (using the objective in (\ref{eq:loss}), the $\mbox{KL}$ term between an unseen class test input $\hat{\xv}_i$ and its class based prior is given by $-\mbox{KL}(q_\phiv(\zv|\hat{\xv}_i)||p_\psiv(\zv|\av)))$. If we had access to the true labels of these inputs, we could add those directly to the original optimization problem ((\ref{eq:loss})). However, since we do not know these labels, we propose an unsupervised method that can still use these unlabeled inputs to \emph{refine} the inductive model presented in the previous section. 

A na\"ive approach for directly leveraging the unlabeled inputs in (\ref{eq:loss}) without their labels would be to add the following reconstruction error term to the objective
\begin{align}\label{eq:transRecons}
	\tilde{\Lcal}_{\theta,\phi,\psi}(\hat{\xv}, \av) &= \E_{q_\phi(\zv|\xv)}[\log p_\theta(\hat{\xv}|\zv)] 
\end{align}

\noindent However, since this objective completely ignores the label information of $\hat{\xv}$, it is not expected to work well in practice and only leads to marginal improvements over the purely inductive case (as corroborated in our experiments). 

To better leverage the unseen class test inputs in the transductive setting, we augment the inductive ZSL objective (\ref{eq:loss}) with an additional unlabeled data based regularizer that uses only the unseen class test inputs. 

This regularizer is motivated by the fact that the inductive model is able to make reasonably confident predictions (as measured by the predicted class distributions for these inputs) for unseen class test inputs, and these confident predicted class distributions can be emphasized in this regularizer to guide those ambiguous test inputs. To elaborate the regularizer, we first define the inductive model's predicted \emph{probability} of assigning an unseen class test input $\hat{\xv}_i$ to class $c \in \{S+1,\ldots,S+U\}$ to be
\begin{align}\label{eq:probxc}
	q(\hat{\xv}_i,c)=\frac{\exp(-\mbox{KL}(q_\phiv(\zv|\hat{\xv}_i)||p_\psiv(\zv|\Amat_c)))}{\sum_{c} \exp(-\mbox{KL}(q_\phiv(\zv|\hat{\xv}_i)||p_\psiv(\zv|\Amat_c)))}
\end{align}

Our proposed regularizer (defined below in (\ref{eq:transKL})) promotes these class probability estimates $q(\hat{\xv}_i,c)$ to be sharper, i.e., the most likely class should dominate the predicted class distribution $q(\hat{\xv}_i,c)$) for the unseen class test input $\hat{\xv}_i$.

Specifically, we define a sharper version of the predicted class probabilities $q(\hat{\xv}_i,c)$ as $p(\hat{\xv}_i,c)=\frac{q(\hat{\xv}_i,c)^2 /  g(c)}{ \sum_{c'}q(\hat{\xv}_i,c')^2 /  g(c')}$, where $g(c) = \sum_{i=1}^{N^\prime} q(\hat{\xv}_i,c)$ is the marginal probability of unseen class $c$. Note that normalizing the probabilities by $g(c)$ prevents large classes from distorting the latent space. 

We then introduce our $\mbox{KL}$ based regularizer that encourages $q(\hat{\xv}_i,c)$ to be close to $p(\hat{\xv}_i,c)$. This can be formalized by defining the sum of the KL divergences between $q(\hat{\xv}_i,c)$ and $p(\hat{\xv}_i,c)$ for all the unseen class test inputs, i.e,
\beq \label{eq:transKL}
\mbox{KL}(P(\hat{\Xmat}) || Q(\hat{\Xmat})) \triangleq \sum_{i=1}^{N'}\sum_{c=S+1}^{S+U}  p(\hat{\xv}_i,c)\log \frac{p(\hat{\xv}_i,c)}{q(\hat{\xv}_i,c)}
\eeq

A similar approach of \emph{sharpening} was recently utilized in the context of learning deep embeddings for clustering problems~\cite{xie2016unsupervised} and data summarization~\cite{wang2016deep}, and is reminiscent of self-training algorithms used in semi-supervised learning~\cite{nigam2000analyzing}.

Intuitively, unseen class test inputs with \emph{sharp} probability estimates will have a more significant impact on the gradient norm of (\ref{eq:transKL}), which in turn leads to improved predictions on the ambiguous test examples (our experimental results corroborate this). Combining (\ref{eq:transRecons}) and (\ref{eq:transKL}), we have the following objective (which we seek to \emph{maximize}) defined exclusively over the unseen class unlabeled inputs

\vspace{-1em}
\beq \label{eq:trans}
U(\hat{\Xmat}) = \sum_{i=1}^{N'}\E_{q_\phi(\zv|\hat{\xv}_i)}[\log p_\thetav(\hat{\xv}_i|\zv)] -\mbox{KL}(P(\hat{\Xmat}) || Q(\hat{\Xmat}))
\eeq 

We finally combine this objective with the original objective ((\ref{eq:loss})) for the inductive setting, which leads to the overall objective $\sum_{n=1}^{N} \hat{\Lcal}_{\theta,\phi,\psi}(\xv_n,\av_n) + U(\hat{\Xmat})$, defined over the seen class labeled training inputs $\{(\xv_n, \av_n)\}_{n=1}^{N}$ and the unseen class unlabeled test inputs $\{\hat{\xv}_i\}_{i=1}^{N^\prime}$.

Under our proposed framework, it is also straightforward to perform few-shot learning~\cite{lake2015human,vinyals2016matching,ravi2017optimization} which refers to the setting when a small number of labeled inputs may also be available for classes $c=S+1,\ldots,S+U$. For these inputs, we can directly optimize (\ref{eq:loss}) on classes $c=S+1,\ldots,S+U$.

\section{Related Work}
\label{sec:relwork}

Several prior methods for zero-shot learning (ZSL) are based on embedding the inputs into a semantic vector space, where nearest-neighbor methods can be applied to find the most likely class, which is represented as a point in the same semantic space~\cite{socher2013zero,norouzi2013zero}. 
Such approaches can largely be categorized into three types: 
($i$) methods that learn the projection from the input space to the semantic space using either a linear regression or a ranking model~\cite{akata2015evaluation,lampert2014attribute}, or using a deep neural network\cite{socher2013zero}; 
($ii$) methods that perform a ``reverse'' projection from the semantic space to the input space\cite{zhang2016learning}, which helps to reduce the \emph{hubness problem} encountered when doing nearest neighbor search at test time~\cite{radovanovic2010hubs}; and
($iii$) methods that learn a shared embedding space for the inputs and the class attributes~\cite{zhang2016zero,changpinyo2016synthesized}.

Another popular approach to ZSL is based on modeling each unseen class as a linear/convex combination of seen classes~\cite{norouzi2013zero}, or of a set of shared ``abstract'' or ``basis'' classes~\cite{romera2015embarrassingly,changpinyo2016synthesized}. Our framework can be seen as a flexible generalization to the latter type of models since the parameters $\Wmat_\mu$ and $\Wmat_\sigma$ defining the latent space distributions are shared by the seen and unseen classes. 

One general issue in ZSL is the \textit{domain shift} problem -- when the seen and unseen classes come from very different domains. Standard ZSL models perform poorly under these situations. However, utilizing some additional unlabeled data from those unseen domains can somewhat alleviates the problem. To this end, \cite{kodirov2015unsupervised} presented a transductive ZSL model which uses a dictionary-learning-based approach for learning unseen-class classifiers. In their approach, the dictionary is adapted to the unseen-class domain using the unlabeled test inputs from unseen classes. Other methods that can leverage unlabeled data include~\cite{fu2015transductive,rohrbach2013transfer,li2015semi,zhao2016zero}. Our model is robust to the \textit{domain shift} problem due to its ability to incorporate unlabeled data from unseen classes.

Somewhat similar to our VAE based approach, recently \cite{kodirov2017semantic} proposed a semantic autoencoder for ZSL. However, their method does not have a proper generative model. Moreover, it assumes each class to be represented as a fixed point and cannot extend to the transductive setting. 

Deep encoder-decoder based models have recently gained much attention for a variety of problems, ranging from image generation~\cite{rezende2016one} and text matching~\cite{shen2017deconvolutional}. A few recent works exploited the idea of applying sematic regularization to the latent embedding spaced shared between encoder and decoder to make it suitable for ZSL tasks~\cite{kodirov2017semantic,tsai2017learning}. However, these methods lack a proper generative model; moreover ($i$) these methods assume each class to be represented as a fixed point, and ($ii$) these methods cannot extend to the transductive setting. Variational autoencoder~(VAE)~\cite{kingma2014auto} offers an elegant probabilistic framework to generate continues samples from a latent gaussian distribution and its supervised extensions~\cite{kingma2014semi} can be used in supervised and semi-supervised tasks. However, supervised/semi-supervised VAE~\cite{kingma2014semi} assumes all classes to be seen at the training time and the label space $p(y)$ to be discrete, which makes it unsuitable for the ZSL setting. In contrast to these methods, our approach is based on a deep generative framework using a supervised variant of VAE, treating each class as a distribution in a latent space. This naturally allows us to handle the intra-class variability. Moreover, the supervised VAE model helps learning highly discriminative representations of the inputs. 

Some other recent works have explored the idea of generative models for zero-shot learning~\cite{li2017zero,verma2017simple}. However, these are primarily based on linear generative models, unlike our model which can learn discriminative and highly nonlinear embeddings of the inputs. In our experiments, we have found this to lead to significant improvements over linear models~\cite{li2017zero,verma2017simple}. 

Deep generative models have also been proposed recently for tasks involving learning from limited supervision, such as one-shot learning~\cite{rezende2016one}. These models are primarily based on feedback and attention mechanisms. However, while the goal of our work is to develop methods to help recognize previously unseen classes, the focus of methods such as ~\cite{rezende2016one} is on tasks such as generation, or learning from a very small number of labeled examples. It will be interesting to combine the expressiveness of such models within the context of ZSL.

\section{Experiments}
We evaluate our framework for ZSL on several benchmark datasets and compare it with a number of state-of-the-art baselines. Specifically, we conduct our experiments on the following datasets: ($i$) Animal with Attributes (AwA)~\cite{lampert2014attribute}; ($ii$) Caltech-UCSD Birds-200-2011 (CUB-200)~\cite{wah2011caltech}; and ($iii$) SUN attribute (SUN)~\cite{patterson2012sun}. For the large-scale dataset (ImageNet), we follow~\cite{fu2016semi}, for which 1000 classes from ILSVRC2012~\cite{russakovsky2015imagenet} are used as seen classes, while 360 non-overlapped classes of ILSVRC2010~\cite{deng2009imagenet} are used as unseen classes. The statistics of these datasets are listed in Table~\ref{Table:datasets}.

\vspace{-0.1cm}
\begin{table}[!htbp]
	\begin{center}
		\scalebox{0.75}{
			\footnotesize{\begin{tabular}{l|c|c  c | c  c }
					\hline
					\multirow{2}{*}{Dataset} &
					\multirow{2}{*}{\# Attribute} & \multicolumn{2}{c}{training(+validation)} & \multicolumn{2}{c}{testing} \\
					& & \# of images & \# of classes & \# of images & \# of classes \\
					\hline \hline
					AwA & 85 &   24,295 & 40 & 6,180 & 10 \\ \hline
					CUB & 312 &  8,855  & 150&  2,933& 50 \\ \hline
					SUN & 102 &  14,140&  707& 200 & 10  \\ \hline 
					ImageNet & 1,000 & 200,000 & 1,000 & 54,000 & 360 \\ \hline
		\end{tabular}}}
		\vspace{-1em}
		\caption{Summary of datasets used in the evaluation}
		\label{Table:datasets}
	\end{center}
	\vspace{-1em}
\end{table} 

In all our experiments, we consider VGG-19 fc7 features \cite{simonyan2014very} as our raw input representation,  which is a 4096-dimensional feature vector. For the semantic space, we adopt the default class attribute features provided for each of these datasets. The only exception is ImageNet, for which the semantic word vector representation is obtained from word2vec embeddings~\cite{mikolov2013distributed} trained on a skip-gram text model on 4.6 million Wikipedia documents. For the reported experiments, we use the standard train/test split for each dataset, as done in the prior work. For hyper-parameter selection, we divide the training set into training and validation set; the validation set is used for hyper-parameter tuning, while setting $\lambda=1$ across all our experiments.  

For the VAE model, a multi-layer perceptron (MLP) is used for both encoder $q_\phiv(\zv | \xv)$ and decoder $p_\thetav(\xv | \zv)$. The encoder and decoder are defined by an MLP with two hidden layers, with $1000$ nodes in each layer. ReLU is used as the nonlinear activation function on each hidden layer and dropout with constant rate $0.8$ is used to avoid overfitting. The latent space $\zv$ was set to be $100$ for small datasets and $500$ for ImageNet. Our results with variance are reported by repeating with 10 runs. Our model is written in Tensorflow and trained on NVIDIA GTX TITAN X with 3072 cores and 11GB global memory.

We compare our method (referred to as VZSL) with a variety of state-of-the-art baselines using VGG-19 fc7 features and specifically we conduct our experiments on the following tasks:
\begin{itemize}
	\item \textbf{Inductive ZSL:} This is the standard ZSL setting where the unseen class latent space distributions are learned using only seen class data. 
	\item \textbf{Transductive ZSL: } In this setting, we also use the unlabeled test data while learning the unseen class latent space distributions. Note that, while this setting has access to more information about the unseen class, it is only through unlabeled data.
	\item \textbf{Few-Shot Learning:} In this setting~\cite{lake2015human,vinyals2016matching,ravi2017optimization}, we also use a small number of labeled examples from each unseen class.
\end{itemize}

In addition, through a visualization experiment (using t-SNE \cite{maaten2008visualizing}), we also illustrate our model's behavior in terms its ability to separate the different classes in the latent space.

\begin{table*}[!htbp]
	\begin{center}
		\scalebox{0.83}{
			\begin{tabular}{l |  c | c |  c | c || c| r}
				\hline 
				Method &  AwA & CUB-200 &SUN & Average & Method & ImageNet\\
				\hline \hline
				{\cite{lampert2014attribute}} & $57.23$ & $-$ & $72.00$ & $-$ & {DeViSE \cite{frome2013devise}} & $12.8$\\
				{ESZSL \cite{romera2015embarrassingly}} & $75.32\pm 2.28$ & $-$ & $82.10\pm 0.32$ & $-$ & {ConSE \cite{norouzi2013zero}} & $15.5$\\
				{MLZSC \cite{bucher2016improving}} & $77.32\pm 1.03$ & $43.29\pm 0.38$ & $84.41\pm 0.71$ & $68.34$ & {AMP \cite{fu2015zero}}& $13.1$\\
				{SDL \cite{zhang2016zero}} & $80.46\pm0.53$ & $42.11\pm 0.55$ & $83.83\pm 0.29$ & $68.80$ &{SS-Voc \cite{fu2016semi}}& $16.8$\\
				{BiDiLEL \cite{wang2016zero}} &  $79.20$ & $46.70$ & $-$ & $-$ & &\\
				{SSE-ReLU \cite{zhang2015zero}} &  $76.33\pm0.83$ & $30.41\pm 0.20$ & $82.50\pm 1.32$ & $63.08$& &\\
				{JFA \cite{zhang2016learning}} & $81.03\pm 0.88$ & $46.48\pm 1.67$ & $84.10\pm 1.51$ & $70.53$& & \\
				{SAE \cite{kodirov2017semantic}} &$83.40$ & $56.60$ & $84.50$ & $74.83$& &\\
				{GFZSL \cite{verma2017simple}} & $80.83$ & $56.53$ & $86.50$ & $74.59$  & &\\
				\hline	\hline
				{\text{VZSL$^{\#}$}} & $84.45\pm 0.74$ & $55.37\pm 0.59$ & $85.75\pm 1.93$ & $74.52$& - & $22.88$\\
				{VZSL} &  $\mathbf{85.28\pm 0.76}$ &$\mathbf{57.42\pm 0.63}$& $\mathbf{86.75\pm 2.02}$ & $\mathbf{76.48}$ & - & $\mathbf{23.08}$
		\end{tabular}}
		\vspace{-0.5em}
		\caption{\small{Top-1 classification accuracy (\%) on AwA, CUB-200, SUN and Top-5 accuracy(\%) on ImageNet under inductive ZSL. VZSL$^{\#}$ denotes our model trained with the reconstruction term from (\ref{eq:loss}) ignored.}
			\label{table:inductive_zsl}	
			\vspace{-2em}			
		}
	\end{center}
\end{table*}

\subsection{Inductive ZSL}
Table~\ref{table:inductive_zsl} shows our results for the inductive ZSL setting. The results of the various baselines are taken from the corresponding papers or reproduced using the publicly available implementations. From Table~\ref{table:inductive_zsl}, we can see that: ($i$) our model performs better than all the baselines, by a reasonable margin on the small-scale datasets; ($ii$) On large-scale datasets, the margin of improvement is even more significant and we outperform the best-performing state-of-the art baseline by a margin of $37.4\%$;  ($iii$) Our model is superior when including the reconstruction term, which shows the effectiveness of the generative model; ($iv$) Even without the reconstruction term, our model is comparable with most of the other baselines. The effectiveness of our model can be attributed to the following aspects. First, as compared to the methods that embed the test inputs in the semantic space and then find the most similar class by doing a Euclidean distance based nearest neighbor search, or methods that are based on constructing unseen class classified using a weighted combination of seen class classifiers~\cite{zhang2015zero}, our model finds the "most probable class" by computing the distance of each test input from \textit{class distributions}. This naturally takes into account the shape (possibly multi-modal) and spread of the class distribution. Second, the reconstruction term in the VAE formulation further strengthens the model. It helps leverage the intrinsic structure of the inputs while projecting them to the latent space. This aspect has been shown to also help other methods such as~\cite{kodirov2017semantic} (which we use as one of the baseline), but the approach in~\cite{kodirov2017semantic} lacks a generative model. This explains the favorable performance of our model as compared to such methods. 

\subsection{Transductive ZSL}
Our next set of experiments consider the transductive setting. 
Table~\ref{table:transductive_zsl} reports our results for the transductive setting, where we compare with various state-of-the-art baselines that are designed to work in the transductive setting. As Table~\ref{table:transductive_zsl} shows, our model again outperforms the other state-of-the-art methods by a significant margin. We observe that the generative framework is able to effectively leverage unlabeled data and significantly improve upon the results of inductive setting. On average, we obtain about $8\%$ better accuracies as compared to the inductive setting. Also note that in some cases, such as CUB-200, the classification accuracies drop significantly once we remove the VAE reconstruction term. A possible explanation to this behavior is that the CUB-200 is a relative difficult dataset with many classes are very similar to each other, and the inductive setting may not achieve very confident predictions on the unseen class examples during the inductive pre-training process. However, adding the reconstruction term back into the model significantly improves the accuracies. Further, compare our entire model with the one having only (\ref{eq:transRecons}) for the unlabeled, there is a margin for about $5\%$ on AwA and CUB-200, which indicates the necessity of introduced $\mbox{KL}$ term on unlabeled data. 

\begin{table*}	
	\begin{center}
		\scalebox{1}{
			\footnotesize{\begin{tabular}{l | c | c| c | r}
					\hline 
					Method  &  AwA & CUB-200 &SUN & Average\\
					\hline \hline
					{SMS \cite{guo2016transductive}} & $78.47$ & $-$ & $82.00$ & $-$ \\
					{ESZSL \cite{romera2015embarrassingly}} & $84.30$ & $-$ & $37.50$ & $-$\\
					{JFA+SP-ZSR \cite{zhang2016learning}} & $88.04\pm 0.69$ & $55.81\pm 1.37$ & $85.35\pm 1.56$ & $77.85$\\
					{SDL \cite{zhang2016zero}} & $92.08\pm 0.14$ & $55.34\pm 0.77$ & $86.12\pm 0.99$ & $76.40$\\
					{DMaP \cite{li2017zero}} &  $85.66$ & $61.79$ & $-$ & $-$\\
					{TASTE \cite{yu2017transductive}} &  $89.74$ & $54.25$ & $-$ & $-$\\
					{TSTD \cite{yu2017transductive2}}  &  $90.30$ & $58.20$ &  $-$ & $-$ \\
					{GFZSL \cite{verma2017simple}} & $94.25$ & $63.66$ & $87.00$ & $80.63$ \\
					\hline	\hline
					{VZSL$^{\#}$}  & $93.49\pm 0.54$ & $59.69\pm 1.22$ & $86.37\pm 1.88$ & $79.85$ \\
					{VZSL$^{\star}$}  & $87.59\pm 0.21$  & $61.44 \pm 0.98$ & $86.66\pm 1.67$ & $77.56$ \\
					{VZSL}  & $\mathbf{94.80\pm 0.17}$ & $\mathbf{66.45\pm 0.88}$ & $\mathbf{87.75\pm 1.43}$ & $\mathbf{83.00}$
		\end{tabular}}}
		\caption{\small{Top-1 classification accuracy (\%) obtained on AwA, CUB-200 and SUN under transductive setting. VZSL$^{\#}$ denotes our model with VAE reconstruction term ignored. VZSL$^{\star}$ denotes our model with only Eq~(\ref{eq:transRecons}) for unlabeled data. The '-' indicates the results was not reported}}
		\label{table:transductive_zsl} 			
	\end{center}
\end{table*}

\subsection{Few-Shot Learning (FSL)}
\begin{figure}[!htbp]
	\begin{center}
		\vspace{-1em}
		\includegraphics[scale=0.39]{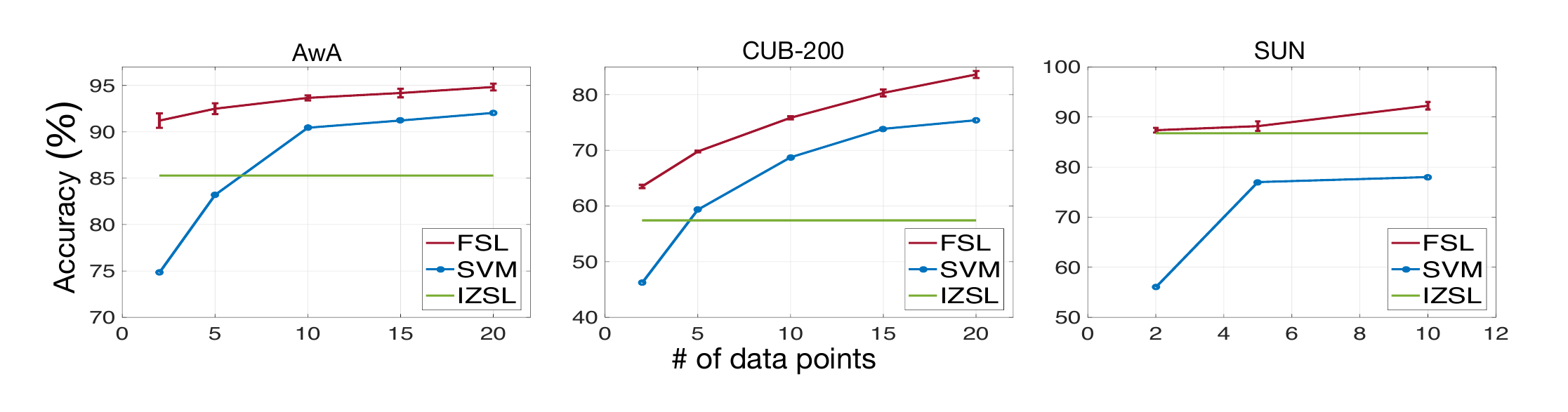}
		\vspace{-2em}
		\caption{\small{Accuracies (\%) in FSL setting: For each data set, results are reported using 2,5,10,15,20 labeled examples for each unseen class}}
		\label{fig:fsl}
	\end{center}
	\vspace{-1em}
\end{figure}

In this section, we report results on the task of FSL~\cite{salakhutdinov2013learning,mensink2014costa} and transductive FSL \cite{frome2013devise} \cite{socher2013zero}. 
In contrast to standard ZSL, FSL allows leveraging a few labeled inputs from the unseen classes, while the transductive FSL additionally also allows leveraging unseen class unlabeled test inputs. To see the effect of knowledge transfer from the seen classes, we use a multiclass SVM as a baseline that is provided the same number of labeled examples from each unseen class. In this setting, we vary the number of labeled examples from 2 to 20 (for SUN, we only use 2, 5 and 10 due to the small number of labeled examples). In Figure~\ref{fig:fsl}, we also compared with standard inductive ZSL which does not have access to the labeled examples from the unseen classes. Our results are shown in Figure~\ref{fig:fsl}. 

As can be seen, even with as few as 2 or 5 additional labeled examples per class, the FSL significantly improves over ZSL. We also observe that the FSL outperform a multiclass SVM which demonstrates the advantage of the knowledge transfer from the seen class data.  Table~\ref{table:transductive_fsl} reports our results for the transductive FSL setting where we compare with other state-of-the-art baselines. In this setting too, our approach outperforms the baselines.

\begin{table}[!htbp]
	\begin{center}
		\caption{\small{Transductive few-shot recognition comparison using top-1 classification accuracy (\%). For each test class, 3 images are randomly labeled, while the rest are unlabeled}}
		\vspace{1em}
		\scalebox{0.78}{
			\begin{tabular}{l | c | c |  r}
				\hline 
				Method & AwA & CUB-200 & Average\\
				\hline \hline
				{DeViSE \cite{frome2013devise}} & $92.60$ & $57.50$ & $75.05$ \\
				{CMT \cite{socher2013zero}}    & $90.60$ & $62.50$ & $76.55$ \\
				{ReViSE \cite{tsai2017learning}} & $94.20$ & $68.40$ & $81.30$\\
				\hline	\hline
				{VZSL} & $\mathbf{95.62\pm 0.24}$ & $\mathbf{68.85\pm 0.69}$ & $\mathbf{82.24}$
				\label{table:transductive_fsl}
		\end{tabular}}
	\end{center}
	\vspace{-2em}
\end{table}
\begin{figure*}[!htbp]
	\begin{center}
		\includegraphics[width=\textwidth]{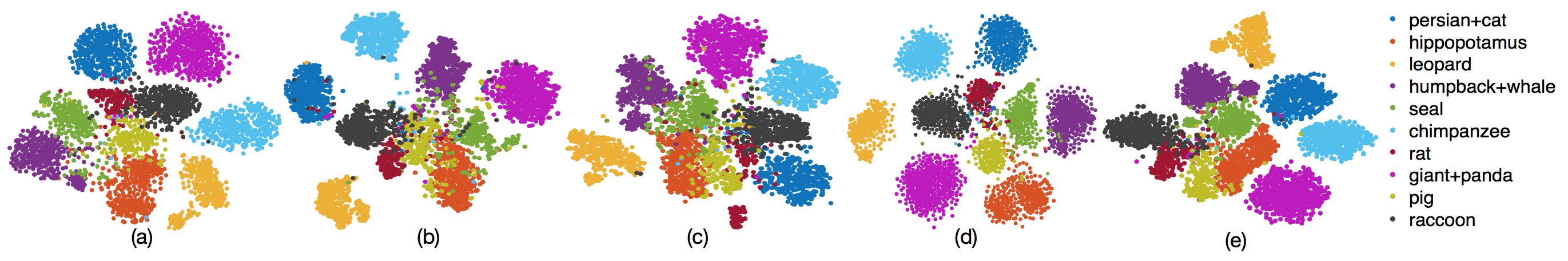}
		\vspace{-2em}
		\caption{\small{t-SNE visualization for AwA dataset (a) Original CNN features (b) Latent code for our VZSL under inductive zero-shot setting (c) Reconstructed features under inductive zero-shot setting (d) Latent code for our VZSL under transductive zero-shot setting (e) Reconstructed features under transductive setting. Different colors indicate different classes.}}
		\label{fig:tSNE_exp4}
	\end{center}
	\vspace{-2em}
\end{figure*}

\subsection{t-SNE Visualization}
To show the model's ability to learn highly discriminative representations in the latent embedding space, we perform a visualization experiment. Figure~\ref{fig:tSNE_exp4} shows the t-SNE \cite{maaten2008visualizing} visualization for the raw inputs, the learn latent embeddings, and the \emph{reconstructed} inputs on AwA dataset, for both inductive ZSL and transductive ZSL setting. 

As can be seen, both the reconstructions and the latent embeddings lead to reasonably separated classes, which indicates that our generative model is able to learn a highly discriminative latent representations. We also observe that the inherent correlation between classes might change after we learn the latent embeddings of the inputs. For example, "giant+panda" is close to "persian+cat" in the original CNN features space but far away from each other in our learned latent space under transductive setting. A possible explanation could be that the sematic features and image features express information from different views and our model learns a representation that is sort of a compromise of these two representations. 

\section{Conclusion}

We have presented a deep generative framework for learning to predict unseen classes, focusing on inductive and transductive zero-shot learning (ZSL). In contrast to most of the existing methods for ZSL, our framework models each seen/unseen class using a class-specific latent-space distribution and also models each input using a VAE-based decoder model. Prediction for the label of a test input from any unseen class is done by matching the VAE posterior distribution for the latent representation of this input with the latent-space distributions of each of the unseen class. This distribution matching method in the latent space provides more robustness as compared to other existing ZSL methods that simply use a point-based Euclidean distance metric. Our VAE based framework leverages the intrinsic structure of the input space through the generative model. Moreover, we naturally extend our model to the transductive setting by introducing an additional regularizer for the unlabeled inputs from unseen classes.  We demonstrate through extensive experiments that our generative framework yields superior classification accuracies as compared to existing ZSL methods, on both inductive ZSL as well as transductive ZSL tasks. The proposed framework can scale up to large datasets and can be trained using any existing stochastic gradient based method. Finally, although we use isotropic Gaussian to model each model each seen/unseen class, it is possible to model with more general Gaussian or any other distribution depending on the data type. We leave this possibility as a direction for future work. \\

\noindent\textbf{Acknowledgements}: This research was supported in part by grants from DARPA, DOE, NSF and ONR.

\clearpage

	\clearpage
	{
	\fontsize{8.5pt}{7.5}\selectfont
	\bibliographystyle{aaai} 
	\bibliography{references/zsl}}

\end{document}